\documentclass[preprint]{elsarticle}

\usepackage{lineno,hyperref}
\hypersetup{
	colorlinks=true,
	linkcolor=blue,
	filecolor=magenta,
	urlcolor=cyan,
	breaklinks=true
}

\usepackage[utf8]{inputenc}
\usepackage{algorithm}
\usepackage{algpseudocode}
\usepackage{pdflscape}
\usepackage{booktabs}
\usepackage{natbib}
\usepackage{subfig}
\usepackage{rotating}
\usepackage{pdflscape}
\usepackage{amssymb, amsmath, amsbsy}
\usepackage[swedish,english]{babel}
\usepackage{etoolbox}
\makeatletter
\providecommand{\subtitle}[1]{
	\apptocmd{\@title}{\par {\large #1 \par}}{}{}
}
\makeatother

\makeatletter
\def\ps@pprintTitle{%
	\let\@oddhead\@empty
	\let\@evenhead\@empty
	\def\@oddfoot{}%
	\let\@evenfoot\@oddfoot}
\makeatother

\journal{Journal of \LaTeX\ Templates}

\algblockdefx{MAP}{ENDMAP}[1]%
{\textbf{map} #1}%
{\textbf{end map}}

\begin{document}

\begin{frontmatter}

\title{Smart Data driven Decision Trees Ensemble Methodology for Imbalanced Big Data}

\author[granada]{Diego Garc\'ia-Gil\corref{mycorrespondingauthor}}
\ead{djgarcia@decsai.ugr.es}

\author[granada]{Salvador Garc\'ia}
\ead{salvagl@decsai.ugr.es}

\author[mdh]{Ning Xiong}
\ead{ning.xiong@mdh.se}

\author[granada]{Francisco Herrera}
\ead{herrera@decsai.ugr.es}

\address[granada]{Department of Computer Science and Artificial Intelligence, Andalusian Research Institute in Data Science and Computational Intelligence (DaSCI), University of Granada, 18071, Granada, Spain}
\address[mdh]{School of Innovation Design, and Engineering, Mälardalen University Västerås, SE-72123, Sweden}
\cortext[mycorrespondingauthor]{Corresponding author}

\begin{abstract}
	
	\paragraph{Background}
	Differences in data size per class, also known as imbalanced data distribution, have become a common problem affecting data quality. Big Data scenarios pose a new challenge to traditional imbalanced classification algorithms, since they are not prepared to work with such amount of data. Split data strategies and lack of data in the minority class due to the use of MapReduce paradigm have posed new challenges for tackling the imbalance between classes in Big Data scenarios. Ensembles have shown to be able to successfully address imbalanced data problems. Smart Data refers to data of enough quality to achieve high performance models. The combination of ensembles and Smart Data, achieved through Big Data preprocessing, should be a great synergy.
	
	\paragraph{Methods}
	In this paper, we propose a novel Smart Data driven Decision Trees Ensemble methodology for addressing the imbalanced classification problem in Big Data domains, namely SD\_DeTE methodology. This methodology is based on the learning of different decision trees using distributed quality data for the ensemble process. This quality data is achieved by fusing Random Discretization, Principal Components Analysis and clustering-based Random Oversampling for obtaining different Smart Data versions of the original data.
	
	\paragraph{Results}
	Experiments carried out in 21 binary adapted datasets have shown that our methodology outperforms Random Forest.

\end{abstract}


\end{frontmatter}


\section{Introduction}

We are experiencing a constant revolution in terms of data generation and transmission speeds. Technologies such as LTE/4G networks have been surpassed by faster standards like the novel 5G network~\cite{morocho2019machine}. This increasing amount of data contains very valuable insights for businesses. This is the era of Big Data~\cite{ramirez2018big}. Big Data can be defined as a high \textit{volume} of data, generated at a high \textit{velocity}, composed of a wide \textit{variety} of data types, with a potential high \textit{value}, and high \textit{veracity}. This conforms what is known as the five Big Data V's (among many others)~\cite{10.1145/3312614.3312623}.

Most of nowadays real-world data is generated from an Internet of Things (IoT) context~\cite{10.1145/3419634}. This IoT scenario is composed of a myriad of sensors that generate temporal data in the form of time series~\cite{GE2018601} or tabular data~\cite{shwartzziv2021tabular}. Real-world classification problems based on tabular data are not usually balanced. This means that one class (usually the one that contains the concept of interest) is underrepresented in the dataset~\cite{THABTAH2020429}. This is known as imbalanced classification~\cite{fernandez2018learning}, and causes machine learning algorithms to bias towards the class with the greater representation. The imbalanced classification task have been extensively researched in the literature~\cite{fernandez2018learning}.

Imbalanced classification has a critical role in Big Data environments, where the imbalance between classes may be greater. This is known as imbalanced Big Data classification~\cite{libro}. Despite the extensive list of imbalanced classification methods proposed in the literature, we can find only a handful of classic sampling proposals extended to Big Data domains, such as Random OverSampling (ROS), Random UnderSampling (RUS)~\cite{Fernandez2017}, or ``Synthetic Minority Oversampling TEchnique'' (SMOTE)~\cite{basgall2018smote, gutierrez2017smote}. As stated by recent surveys~\cite{leevy2018survey, juez2021experimental}, current imbalanced Big Data proposals are usually a direct extension of classic oversampling imbalanced methods to Big Data environments. This entails a key issue for those methods, which is suffering from lack of data in the different maps within the already very small minority class space~\cite{leevy2018survey}. SMOTE and its extensions constitute the current state-of-the-art for imbalanced problems, however, it lacks a quality extension to Big Data environments due to suffering from the aforementioned issue~\cite{juez2021experimental}. This leads to a sub-par performance of these methods in Big Data domains~\cite{leevy2018survey}.

In imbalanced Big Data scenarios there is the challenge of new approaches that take into account the peculiarities of distributed MapReduce processing and the availability of several \textit{maps} with imbalanced subsets that require their own processing~\cite{leevy2018survey}. In~\cite{juez2021experimental}, authors identified two gaps within the imbalanced Big Data classification scenario: \textit{``the few ensemble methods designed for Big Data problems, and perhaps even fewer for processing imbalance within Big Data''}. This assessment for the design of efficient distributed algorithms, in particular ensembles, capable of analyzing the nature of \textit{maps} by performing an imbalanced analysis of the data~\cite{libro}, drive our current proposal, advancing towards the use of Smart Data and ensembles.

Recently, the term Smart Data has emerged in the Big Data ecosystem. Smart Data refers to the challenge of extracting quality data from raw Big Data~\cite{ramirez2018big, libro}. This new concept aims to achieve quality data with \textit{value} and \textit{veracity} properties~\cite{XIE2021168}. Data preprocessing clearly resembles the concept of Smart Data to ensure achieving quality data. Data preprocessing is also inherent in all imbalanced approaches~\cite{fernandez2018learning}. In Big Data environments, Big Data preprocessing has a crucial role for enabling Smart Data~\cite{libro}. On the other hand, ensembles have established as the most popular algorithm-level solution for tackling the imbalanced classification problem~\cite{fernandez2018learning, galar2011review}. Ensembles and Smart Data have proven to perform consistently in Big Data environments when facing label noise~\cite{GARCIAGIL2019135, doi:10.1002/int.22193}. Our hypothesis in this paper is their combined use to tackle the imbalanced Big Data classification problem.

We propose a novel Smart Data driven Decision Trees Ensemble methodology for addressing the imbalanced Big Data classification, namely SD\_DeTE methodology. SD\_DeTE methodology produces a decision tree-based ensemble combined with Smart Data for introducing diversity in the datasets, creating different decision trees that result in an efficient distributed ensemble algorithm. Quality data is achieved through the application of several data preprocessing techniques in order to enable different Smart Data approaches of the dataset, that will enable the learning of better base classifiers and to achieve efficient distributed algorithms. Therefore, SD\_DeTE methodology is composed of a Smart Data generation process, and an ensemble learning process:

\begin{enumerate}
	\item \textit{Smart Data}: the first objective is to add the required level of diversity to the dataset. For this, the combination of Random Discretization (RD) and randomized Principal Component Analysis (PCA), proposed in Principal Components Analysis Random Discretization Ensemble (PCARDE) algorithm~\cite{GARCIAGIL2018}, is used. For a data balancing step, a novel combination of clustering and ROS is presented. SD\_DeTE methodology performs clustering to the expanded data resulting from the combination of RD and PCA datasets. Then, it balances the clusters using ROS technique. The result of this process is a distributed Smart Data version of the dataset, with the appropriate level of diversity.
    
    \item \textit{Ensemble Learning}: this process creates the ensemble through the learning of different base classifiers using a decision tree as a classifier. The distributed Smart Data will produce better base classifiers.
\end{enumerate}

To assess the performance of SD\_DeTE methodology, we have conducted an extensive experimentation, using 21 binary adapted Big Data imbalanced datasets. All datasets have been selected from the latest literature in tabular data and Big Data. We have compared SD\_DeTE methodology against Spark's MLlib implementation of a decision tree, Random Forest~\cite{spark2016}, and PCARDE algorithm~\cite{GARCIAGIL2018}. These three classifiers have been tested without any data balancing technique applied, and using RUS, ROS and SMOTE techniques. Results obtained have been validated by different Bayesian Sign Tests, in order to assess if SD\_DeTE methodology achieves statistically better performance than the rest of the tested methods ~\cite{10.1007/978-3-319-59650-1_24}.

The rest of this paper is organized as follows: Section~\ref{sec:background} gives a description of the imbalanced data classification and Big Data problem. Section~\ref{sec:proposal} describes the proposal in detail. Section~\ref{sec:experiments} shows all the experiments carried out to prove the performance of SD\_DeTE methodology for several Big Data problems. Finally, Section~\ref{sec:conclusions} concludes the paper.

\section{Related work}
\label{sec:background}

In this section, we provide an introduction to the class imbalance problem in classification, among with the different proposals to tackle it (Section~\ref{sec:imb}). Then, the state of Big Data and MapReduce framework is analysed in Section~\ref{sec:mapreduce}. The state-of-the-art regarding imbalanced Big Data scenario is depicted in Section~\ref{sec:imbbd}.

\subsection{Imbalanced Data Classification}
\label{sec:imb}

In a binary classification problem, a dataset is said to be imbalanced when there is a notable difference in the number of instances belonging to different classes~\cite{fernandez2018learning}. The class with the larger number of instances is known as the majority class. Similarly, the class with the lower number of instances is known as the minority class, and usually contains the concept of interest.

As stated earlier, this problem poses a major challenge to standard classifier learning algorithms, since they will bias towards the class with the greater representation, as their internal search process is guided by a global search measure weighted in favor of accuracy~\cite{libro}. In datasets with high imbalance ratio (IR), classifiers that maximize the accuracy will treat the minority class as noise and ignore it, achieving a high accuracy by only classifying the majority class, since more general rules will be preferred.

Many techniques have been proposed to tackle imbalanced data classification. However, ensembles have established themselves as the state-of-the-art in performance~\cite{fernandez2018learning, BI201881, galar2011review}. Because of their accuracy orientation, ensembles cannot be directly applied to imbalanced datasets, since the base classifiers will ignore the minority class. Their combination with other techniques that tackle the class imbalance problem can improve ensemble performance in these scenarios. These hybrid approaches involve the addition of a data sampling step that allows the classifier to better detect the different classes.

In the literature, data preprocessing methods for imbalanced data classification can be divided into different categories: oversampling methods, undersampling methods, and hybrid approaches~\cite{libro, fernandez2018learning}. The former (such as ROS~\cite{Batista:2004:SBS:1007730.1007735}) replicates the minority class instances until a certain balance is reached. On the other hand, undersampling techniques (such as RUS~\cite{Batista:2004:SBS:1007730.1007735}) remove examples from the majority class until the proportion of classes is adjusted. Hybrid approaches combine the previous two techniques, usually starting with an oversampling of the data, followed by an undersampling step that removes samples from both classes, in order to remove noisy instances and improve the classifier performance.

The SMOTE algorithm, along with its many extensions~\cite{gutierrez2017smote, fernandez2018smote}, constitute the current state-of-the-art in data preprocessing for imbalanced data. It adds synthetic instances from the minority class until the class distribution is balanced. Those new instances are created by the interpolation of several minority class instances that belong to the same neighborhood. SMOTE calculates the \textit{k} nearest neighbors of each minority class example. Then, in the segment that connects every instance with its \textit{k} closest neighbors, a synthetic instance is randomly created~\cite{ma2019predicting}.

Clustering has also been employed effectively for the data imbalanced problem as a way to increase the density of points belonging to certain neighborhoods~\cite{nejatian2018using, LE2021107033}. These methods balance the data by localizing groups of instances belonging to different neighborhoods, and then applying a data sampling technique, improving the later learning process~\cite{lin2017clustering, zhang2010cluster}.

Performance evaluation is a key factor for assessing the classification performance. In binary classification problems, the confusion matrix (shown in Table~\ref{tab:conf}) collects correctly and incorrectly classified examples from both classes.

\begin{table}[htb!]
	\centering
	\caption{Confusion Matrix for Binary Classification Problems}
	\label{tab:conf}
	\begin{tabular}{lll}
		\toprule
		                &   Positive Prediction &   Negative Prediction \\
		\midrule
        Positive class  &   True Positive (TP)  &   False Negative (FN) \\
        Negative class  &   False Positive (FP) &   True Negative (TN)  \\
		\bottomrule 
	\end{tabular}
\end{table}

Traditionally, accuracy (Equation~\ref{eq:acc}) has been the most extended and widely used metric for assessing classification performance. However, accuracy is not a valid metric when dealing with imbalanced datasets, since it will not show the classification of both classes, only the majority class, and it will led to wrong conclusions.

\begin{equation}\label{eq:acc}
    Acc = \dfrac{TP + TN}{TP + FN + FP + TN}
\end{equation}

The Geometric Mean (GM), described in Equation~\ref{eq:gm}, attempts to maximize the accuracy of both minority and majority classes at the same time~\cite{BARANDELA2003849}. The accuracy of both minority and majority classes is represented by the True Positive Rate (TPR) $= \frac{TP}{TP+FN}$ and True Negative Rate (TNR) $= \frac{TN}{TN+FP}$.

\begin{equation}\label{eq:gm}
    GM = \sqrt{TPR*TNR}
\end{equation}

Another popular evaluation metric for imbalanced data is the Are Under the Curve (AUC)~\cite{huang2005using}. AUC combines the classification performance of both classes, showing the trade-off between the TPR and False Positive Rate. This metric provides a single measure of a classifier performance, compared against a random classifier.

\subsection{Big Data \& MapReduce}
\label{sec:mapreduce}

In order to tackle Big Data problems, not only new algorithms are needed, but also new frameworks that operate in distributed clusters are required. Google introduced MapReduce paradigm in 2004~\cite{mapreduce}. This paradigm is nowadays the most popular and widely used paradigm for Big Data processing. It was born for allowing users to generate and/or process Big Data problems, while minimizing disk and network use.

MapReduce follows the simple but powerful divide and conquer approach. It can be divided in two phases, the \textit{map} and \textit{reduce} phase. Before entering the \textit{map} stage, all data is splitted and distributed across the cluster by the master node. The \textit{map} function applies a transformation to each key-value pair located in each computing node. This way, all data is processed independently in a distributed fashion. When the \textit{map} phase is finished, all pairs of data belonging to the same key are redistributed across the cluster. Once all pairs belonging to the same key are located in the same computing node, the \textit{reduce} stage begins. The \textit{reduce} phase can be seen as a aggregation operation that generates the final values.

MapReduce is a programming paradigm for dealing with Big Data. Apache Hadoop is the most popular open-source implementation of the MapReduce paradigm~\cite{White:2012:HDG:2285539}. Despite its popularity and performance, Hadoop presents some important limitations~\cite{DBLP:journals/corr/abs-1209-2191}:
\begin{itemize}
    \item Not suitable for iterative algorithms.
    \item Very intensive disk usage. All \textit{map} and \textit{reduce} processes are read/write from/to disk.
    \item No in-memory computation.
\end{itemize}

Apache Spark can be seen as the natural evolution of Hadoop. It is an open-source framework, focused on speed, easy of use, and advanced analytics~\cite{spark}. Spark is the solution of Hadoop problems, it has in-memory computation, and allows in-memory data persistence for iterative processes. Spark is built on top of a novel distributed data structure, namely Resilient Distributed Datasets (RDDs)~\cite{Zaharia:2012:RDD:2228298.2228301}. These data structures are immutable and unsorted by nature. They can be persisted in memory for repetitive uses, and tracked using a lineage, so that each split can be computed again in case of data lost. RDDs support two types of operations: \textit{transformations} and \textit{actions}. The former transforms the dataset by applying a function to each split, and produces a new RDD. They are lazy operations, meaning that they are not computed until needed. On the other hand, \textit{actions} triggers all previous \textit{transformations} of an RDD, and return a value.

In 2012, a distributed machine learning library was created as an extra component of Apache Spark, named MLlib~\cite{meng2016mllib}. It was released and open-sourced to the community in 2013. The number of contributions has been increasing steadily since its conception, making it the most popular machine learning library for Big Data processing nowadays. MLlib includes several algorithms for alike tasks, such as: classification, clustering, regression, or data preprocessing.

\subsection{Imbalanced Big Data}
\label{sec:imbbd}

With the automation in data acquisition and storage, and the explosion of sensors and available data, the problem of imbalanced data classification has been severely affected. It is considered to be one of the worsened or even directly provoked problems by Big Data~\cite{libro}. Moreover, classic algorithms are not able to tackle the imbalanced problem in a reasonable amount of time.

The imbalanced data classification problem has not been disregarded in Big Data. However, most of the proposals in the literature follow two main approaches: data sampling, and distance-based solutions. The former consists of an adaptation of ROS and RUS methods to Big Data domains using the MapReduce paradigm~\cite{DELRIO2014112}. Distance-based methods are mainly composed of different adaptations of SMOTE algorithm to Big Data environments, ranging from an exact version of SMOTE~\cite{basgall2018smote}, SMOTE for multi-class problems~\cite{SLEEMANIV2021106598}, or a GPU-based SMOTE~\cite{gutierrez2017smote}. There are also proposals that combine both approaches in the form of an ensemble~\cite{zhai2017classification}. However, SMOTE algorithm for Big Data scenarios is affected by the lack of data in the different maps, and the presence of small disjuncts~\cite{leevy2018survey, juez2021experimental}.

In recent surveys~\cite{leevy2018survey, juez2021experimental, libro}, authors agree that, in comparison with studies for standard problems, there is still little research devoted to address the problem of imbalanced classification in Big Data scenarios. There is a need for proposals born for and to tackle imbalanced Big Data problems effectively and efficiently. In particular, the design and implementation of new classifiers for Big Data frameworks, capable of internally process the imbalanced situation, are of special interest~\cite{juez2021experimental}. The thorough design at the implementation level of algorithms to address imbalanced Big Data problems is one of the open challenges nowadays~\cite{libro}. Therefore, we aim to provide an efficient and effective ensemble methodology design for the classification of imbalanced Big Data problems.

\section{An Smart Data driven Decision Trees Ensemble Methodology for Imbalanced Big Data}
\label{sec:proposal}

In this section, we describe in detail the proposed ensemble methodology for imbalanced Big Data classification based on achieving diversity and quality data through data preprocessing methods together with decision trees to create the ensemble, SD\_DeTE methodology. It has been designed under the distributed computing paradigm MapReduce, and has been implemented for the Big Data framework Apache Spark~\cite{spark}, which is an extension of such paradigm, making it able to tackle Big Data problems efficiently. SD\_DeTE methodology is available publicly as a Spark package in Spark's third party repository Spark Packages\footnote{\url{https://spark-packages.org/package/djgarcia/Imbalanced-Classification-Ensemble}}.

In Section~\ref{sec:dete_sd}, we explain the details of the Smart Data generation process of SD\_DeTE methodology. Section~\ref{sec:dete_ensemble} details the ensemble learning process. Section~\ref{sec:primitives} describes the Spark primitives used for the implementation of the proposal. Finally, Section~\ref{sec:implementation} depicts the implementation details of the methodology.

\subsection{SD\_DeTE Methodology: Smart Data}
\label{sec:dete_sd}

This ensemble classifier for imbalanced Big Data problems is based on the creation of \textit{smart} datasets for improving the performance of the models learned from the different base classifiers. Diversity is key when working with ensembles. Diversity can be introduced
through small changes in input data, or small changes in the parameters of the classifier. With diversity in the base classifiers, ensembles will be more robust to noise and outliers, and will achieve better performance~\cite{GARCIAGIL2018}. SD\_DeTE methodology achieves a Smart Data version of the dataset with the appropriate level of diversity by using the following two modules:

\paragraph{RD-PCA Module} SD\_DeTE methodology achieves the required diversity by the use of several randomized data preprocesing methods, such as RD and PCA. RD method~\cite{6574846} discretizes the data in $cuts$ intervals by randomly selecting $cuts-1$ instances. Those selected values are sorted and used as thresholds for the discretization of each feature. This mechanism enables RD to produce diversity efficiently each time it is performed on such data. On the other hand, PCA selects a number of variables in a dataset, whilst retaining as much of the variation present in the dataset as possible. This selection is achieved by finding the combinations of the original features to produce principal components, which are uncorrelated. PCA always produces the same result for a fixed number of principal components. In order to achieve the required level of diversity, a random number of selected components is used. The number of components must be in the interval $[1, T-1]$, $T$ being the total number of features of the input data.

Both RD and PCA methods are applied to the input data. Then, the resulting datasets of RD and PCA are joined together feature-wise. This data is a more informative version of the dataset with the appropriate level of diversity, as demonstrated in~\cite{GARCIAGIL2018}. Such dataset needs to be balanced in order to correctly identify the minority and majority classes.

\paragraph{C-ROS Module} A novel combination of hierarchical clustering and oversampling is proposed. Bisecting k-Means is a hierarchical clustering method that uses a divisive (or ``top-down'') approach~\cite{steinbach2000comparison}. The algorithm starts from a single cluster that contains all points. Iteratively it finds divisible clusters on the bottom level and bisects each of them into two clusters using k-Means, until there are $k$ leaf clusters in total or no leaf clusters are divisible. It has been chosen taking into account that it can often be much faster than regular k-Means. Bisecting k-Means has a linear time complexity. In case of a large number of clusters, Bisecting k-Means is even more efficient than k-Means since there is no need to compare every point to each clusters centroid. It just needs to consider the points in the cluster and their distances to two centroids.

Bisecting k-Means is applied to the resulting data from the join of RD and PCA for finding a random number of neighborhoods with a specified maximum of desired clusters. Found clusters are individually balanced using ROS technique until an IR of 1 is reached. The result of this process is a balanced and \textit{smart} dataset with the required level of diversity, which will improve the later learning process by enabling the ensemble to produce efficient distributed algorithms.

\begin{figure}[t!]
	\centering
	\hspace{-1.5cm}
	\includegraphics[scale=0.56]{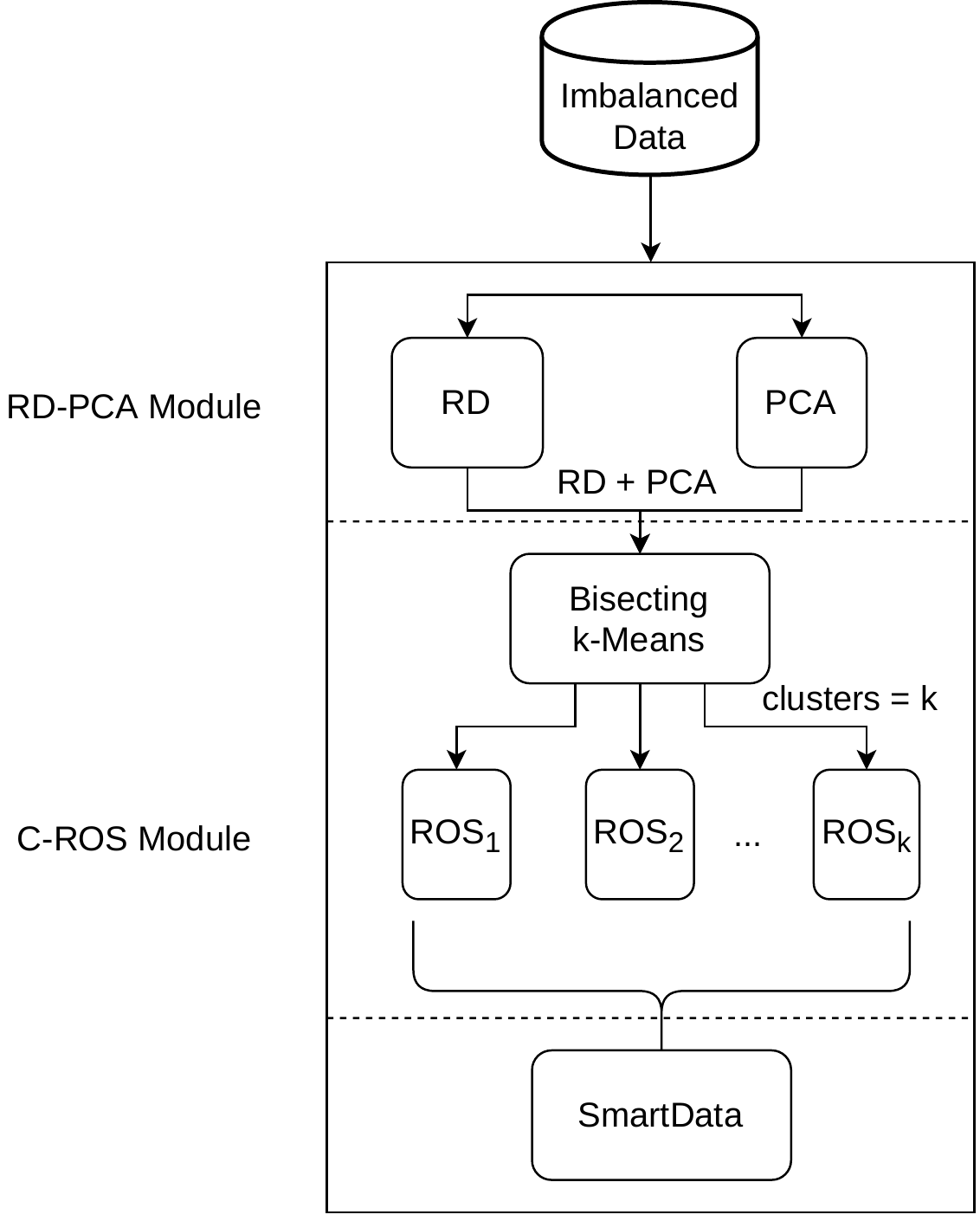}
	\caption{SD\_DeTE methodology Smart Data generation flowchart}
	\label{fig:sd}
\end{figure}

In Figure~\ref{fig:sd} we can see a graphic representation of the Smart Data generation workflow of SD\_DeTE methodology.

\subsection{SD\_DeTE Methodology: Ensemble}
\label{sec:dete_ensemble}

The two previous modules produce a \textit{smart} version of the dataset with the appropriate level of diversity. As stated earlier, ensembles are the most popular solution for tackling imbalanced problems. SD\_DeTE methodology uses the previously generated Smart Data for learning different quality base detectors that will produce a better ensemble method.

\paragraph{Learning Module} Using the previously acquired balanced and \textit{smart} dataset, a decision tree is learned. This decision tree performs a recursive binary partitioning of the input features space. The tree predicts the same label for each leaf partition. These partitions are chosen in a greedy manner, selecting the best split from the set of possible splits, maximizing the information gain at the tree node~\cite{10.5555/2755359}.

\begin{figure}[t!]
	\centering
	\hspace{-1.5cm}
	\includegraphics[scale=0.60]{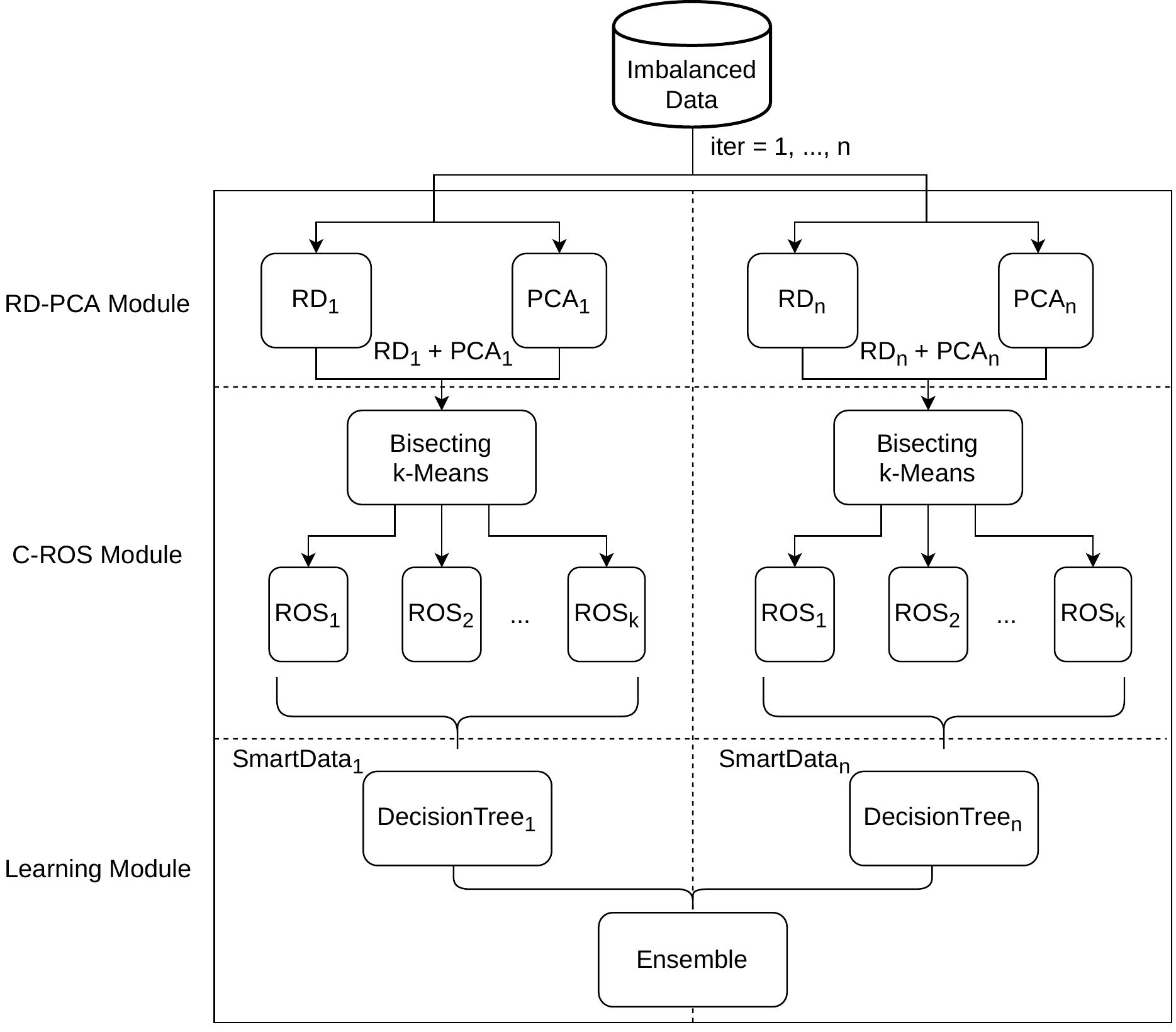}
	\caption{SD\_DeTE methodology learning flowchart}
	\label{fig:ice}
\end{figure}

SD\_DeTE methodology preprocessing and learning process is repeated $iter$ times. In Figure~\ref{fig:ice} we can see a graphic representation of the learning workflow of SD\_DeTE methodology.

All previous steps constitute the learning phase of the ensemble. This phase is composed of $iter$ sub-models, each of them containing the thresholds for RD and the weight matrices for PCA. For the prediction phase of the ensemble, for each data point, the same data preprocessing must be applied. First, data is discretized using the same cut points from RD calculated previously. Then, for selecting the same components as the learning phase, the same weight matrix obtained earlier for PCA at a given iteration is applied to the data. Next, the score of each class is predicted according to the decision tree. This score is calculated by the division of the instances at a leaf node, by the total number of instances. This process is repeated $iter$ times, adding those scores for each instance and iteration. Once this process is finished, for each instance, the class with the largest score is selected as the decision of the ensemble.

\subsection{Spark Primitives}
\label{sec:primitives}

For the implementation of the ensemble, some basic Spark primitives have been used. Here we outline those more relevant for the ensemble~\footnote{For a complete description of Spark’s operations, please refer to Spark’s API: \url{http://spark.apache.org/docs/latest/api/scala/index.html}}:

\begin{itemize}
	\item \textit{map}: applies a transformation to each element of an RDD. Once that transformation has been applied, it returns a new RDD.
	\item \textit{union}: merges two RDDs instance-wise and returns a new RDD.
	\item \textit{zip}: zips two RDDs together.
	\item \textit{filter}: selects all the instances in an RDD that satisfy a condition as a new RDD.
\end{itemize}

These Spark primitives from Spark API are used in the following section, where the implementation of SD\_DeTE methodology is described.

\subsection{SD\_DeTE Methodology Implementation Details}
\label{sec:implementation}

This section describes all the implementation details of SD\_DeTE methodology. Both learning and prediction phases are implemented under Apache Spark, following the MapReduce paradigm.

\subsubsection*{Ensemble Learning Phase}
\label{sec:learning}

\begin{algorithm}[t!]
	\floatname{algorithm}{Algorithm}
	\caption{SD\_DeTE methodology learning algorithm}
	\label{alg:SDE}
	\begin{algorithmic}[1]
		\State \textbf{Input:} \textit{data} an RDD of type LabeledPoint (features, label).
		\State \textbf{Input:} \textit{iter} the number of iterations of the ensemble.
		\State \textbf{Input:} \textit{cuts} the number of intervals for the discretization.
		\State \textbf{Input:} \textit{maxClust} the maximum number of clusters.
		\State \textbf{Output:} The model created, an object of class DeTE\_model.
		\For{$i=0...iter$}
		\State \textbf{Random Discretization}
		\State $thresholds(i) \gets compute\_RD\_thresholds(data, cuts)$
		\State $rdData \gets$ \MAP $inst \in data$
		\For{$j=0...length(inst)-1$}
		\State $inst \gets discretize(inst(j), thresholds(i)(j))$
		\EndFor
		\ENDMAP
		\State \textbf{PCA}
		\State $components \gets random(1, length(data)-1)$
		\State $pcaModels(i) \gets PCA(data, components)$
		\State $pcaData \gets transform(data, pcaModels(i))$
		\State $joinedData \gets zip(rdData, pcaData)$
		\State \textbf{Clustering}
		\State $k \gets random(1, maxClust)$
		\State $clustModel \gets hierarchicalClustering(joinedData, k)$
		\State $clustData \gets predict(joinedData, clustModel)$
		\State \textbf{Data Balancing}
		\State $smartData = \emptyset$
		\For{$l=0...k$}
		\State $rosData \gets ROS(filter(clustData, "cluster" = l), 1.0)$
		\State $smartData = union(rosData, smartData)$
		\EndFor
		\State \textbf{Classifier Learning}
		\State $trees(i) \gets decisionTree(smartData)$
		\EndFor
		\State $return(DeTE\_model(iter, thresholds, pcaModels, trees))$
	\end{algorithmic}
\end{algorithm}

Algorithm~\ref{alg:SDE} explains the ensemble learning phase of SD\_DeTE methodology. This process is divided into five steps: RD and PCA calculation in order to introduce diversity to the dataset, cluster search for the discovery of neighborhoods, cluster balancing, and classifier learning.

\paragraph{Step 1} As stated earlier, SD\_DeTE methodology starts by discretizing the training data using RD method (lines 8-14). This is performed through the random selection of $cuts-1$ instances (line 8). Those thresholds are used to discretize the training data using a $map$ function (lines 10-14). For every instance, we assign the corresponding discretized value to each instance's attribute (lines 11-13).

\paragraph{Step 2} Once RD has been applied to the training data, PCA is performed to select randomly the best principal components (lines 16-19). First, a random number of components is selected in the interval $[1, T-1]$ ($T$ being the total number of features of the training data) (line 16). Then, PCA is calculated on the training data, and the best $components$ are selected (lines 17-18). Finally, the resulting data from RD and PCA are joined together feature-wise using a distributed $zip$ function (line 19).

\paragraph{Step 3} With the desired level of diversity added to the dataset, the next step is the hierarchical clustering search (lines 21-23). We have used Spark's MLlib distributed implementation of Bisecting k-Means. First, we select a random number of clusters, with a maximum of $maxClust$ (line 21). Then, clusters are calculated using the previously RD and PCA zipped data (line 22). Once that process is finished, the same zipped data is predicted in order to assign a cluster to each data point (line 23). The prediction is done level-by-level from the root node to a leaf node, and at each node among its children the closest to the input point is selected.

\paragraph{Step 4} Data balancing is applied to each individual cluster found. We apply ROS technique to the minority class of each cluster until both minority and majority classes are equal (lines 25-29). First, an empty set is created for the allocation of the future new dataset (line 25). For each cluster, ROS is applied with an IR of 1 (line 27). That balanced and \textit{smart} data is added to the empty set (line 28).

\paragraph{Step 5} Finally, a decision tree is learned using this \textit{smart} and balanced dataset (line 31). This data preprocessing and learning process is repeated $iter$ times, keeping each iteration, the computed thresholds for RD, the PCA weight matrices, and the learned tree model. Once all trees have been learned, the model is created and returned.

The following input parameters are required: the dataset (\textit{data}), the number of iterations of the ensemble (\textit{iter}), the number of intervals for the discretization (\textit{cuts}), and the maximum number of clusters (\textit{maxClust}).

\subsubsection*{Ensemble Prediction Phase}
\label{sec:prediction}

The ensemble prediction phase is depicted in Algorithm~\ref{alg:SDEModel}. This process is faster than learning, since clustering and data balancing are not required for prediction. Only the application of RD and PCA is required, both using the same models obtained in the ensemble learning phase. This phase is divided in five steps:

\paragraph{Step 1} First, the data point is discretized using the same cut points from the learning phase (lines 10-13).

\paragraph{Step 2} Next, the principal components are calculated using the learning phase weight matrix for that particular iteration (line 15).

\paragraph{Step 3} The next step is to join both RD and PCA results feature-wise using a distributed $zip$ function (line 17). The result is an expanded dataset with the features of both RD and PCA.

\paragraph{Step 4} Prediction is made for the data point using the decision tree learned in that particular iteration of the ensemble (line 19). The scores of each of the $iter$ predictors are added.

\paragraph{Step 5} Once the instance has all $iter$ scores, the class with the largest weight is selected as the decision of the ensemble and returned (lines 22-23).

\begin{algorithm}[htb!]
	\floatname{algorithm}{Algorithm}
	\caption{DeTE\_model prediction algorithm}
	\label{alg:SDEModel}
	\begin{algorithmic}[1]
		\State \textbf{Input:} \textit{iter} the number of iterations of the ensemble.
		\State \textbf{Input:} \textit{cuts} the cut points for the discretization.
		\State \textbf{Input:} \textit{pcaModels} the models for performing PCA.
		\State \textbf{Input:} \textit{trees} the models of the learned trees.
		\State \textbf{Output:} The label of the test data point.
		\Function{predict}{$test: LabeledPoint$}
		\State $scorePredictions \gets \emptyset$
		\For{$i=0...iter$}
		\State \textbf{Random Discretization}
		\State $rdData \gets \emptyset$
		\For{$j=0...length(test)-1$}
		\State $rdData(c) \gets discretize(test(j), cuts(i)(j))$
		\EndFor
		\State \textbf{PCA}
		\State $pcaData \gets transform(test, pcaModels(i))$
		\State \textbf{Data Join}
		\State $joinedData \gets zip(rdData, pcaData)$
		\State \textbf{Prediction}
		\State $scorePredictions \gets scorePredictions + predict(joinedData, trees(i))$
		\EndFor
		\State \textbf{Scoring}
		\State $label \gets indexOfMax(scorePredictions)$
		\State $return(label)$
		\EndFunction
	\end{algorithmic}
\end{algorithm}

\section{Experimental Results}
\label{sec:experiments}

In this section, we describe the experimental study carried out to compare the performance of different approaches to deal with imbalanced Big Data problems against our ensemble methodology proposal. In Section~\ref{sec:setup}, we show a description of all datasets employed in the comparison, followed by the performance metrics and parameters of the algorithms used. All hardware and software resources used to carry out the experimental study are also detailed. We detail the results of the performance metrics and analyze them using statistical tests in Section~\ref{sec:results}. Section~\ref{sec:times} is devoted to the computing times of SD\_DeTE methodology. Finally, we have conducted an additional experiment for showing the performance improvement achieved by our proposed clustering-based ROS technique in Section~\ref{sec:cro}.

\subsection{Experimental Setup}
\label{sec:setup}

We have selected a wide spectrum of Big Datasets for assessing the performance of SD\_DeTE methodology. These datasets have very different properties among them that will allow us to measure the performance and balancing capabilities of our proposal. Specifically, we have selected the Poker Hand dataset, the Record Linkage Comparison Patterns (RLCP), SUperSYmmetric particles (SUSY) and Higgs bosons (HIGGS) datasets~\cite{Baldi:2014kfa}, and the KDD Cup 1999 dataset, a dataset used for the Third International Knowledge Discovery and Data Mining Tools Competition. These binary adapted datasets have been extracted from the UCI Machine Learning Repository~\cite{Dua:2019}, and have been chosen attending to their size, making them suitable for Big Data scenarios and, therefore, unsuitable for iterative processing. We have also selected a real-world imbalanced dataset, the ECBDL14 dataset~\cite{triguero2015rosefw}. ECBDL14 dataset was used as a reference at the ML competition of the Evolutionary Computation for Big Data and Big Learning, under the international conference GECCO-2014. It is a highly imbalanced binary classification dataset, composed of 98\% of negative instances. For this problem, we have used two subsets with the same IR and the best 90 features found in the competition~\cite{triguero2015rosefw}.

Since some of the selected datasets have more than two classes, we have sampled binary datasets from them to address each case separately. In particular, we have selected new datasets using the majority classes against the minority classes. Table~\ref{tab:datasets} shows all the details of the datasets, including the number of instances (\#Inst.), number of attributes (\#Atts.), class distribution and IR.

\begin{table}[htb!]
	\centering
	\caption{Datasets used in the analysis}
	\label{tab:datasets}
	\begin{tabular}{lrccr}
		\toprule
		Dataset & \#Inst. & \#Atts. & \%Class(maj; min) & IR \\
		\midrule
        poker0\_vs\_2 & 450,022 & 10 & (91.32; 8.68) & 10.52  \\
        poker0\_vs\_3 & 428,464 & 10 & (95.99; 4.01) & 23.94 \\
        poker0\_vs\_4 & 414,032 & 10 & (99.23; 0.77) & 128.06 \\
        poker0\_vs\_5 & 412,600 & 10 & (99.60; 0.40) & 250.59 \\
        poker0\_vs\_6 & 411,990 & 10 & (99.70; 0.30) & 337.81 \\
        poker1\_vs\_2 & 385,842 & 10 & (89.89; 10.11) & 8.89 \\
        poker1\_vs\_3 & 363,932 & 10 & (95.24; 4.76) & 20.03 \\
        poker1\_vs\_4 & 349,891 & 10 & (99.11; 0.89) & 110.82 \\
        poker1\_vs\_5 & 347,695 & 10 & (99.55; 0.45) & 221.17 \\
        poker1\_vs\_6 & 347,867 & 10 & (99.68; 0.32) & 308.77 \\
        rlcp & 4,599,153 & 2 & (99.63; 0.37) & 271.12 \\
        susy\_ir4 & 2,712,173 & 18 & (80.00; 20.00) & 4.00 \\
        susy\_ir8 & 2,440,956 & 18 & (88.89; 11.11) & 7.99 \\
        susy\_ir16 & 2,305,347 & 18 & (94.12; 5.88) & 15.99 \\
        higgs\_ir4 & 5,829,123 & 28 & (80.00; 20.00) & 3.99 \\
        higgs\_ir8 & 5,246,211 & 28 & (88.89; 11.11) & 8.00 \\
        higgs\_ir16 & 4,954,752 & 28 & (94.12; 5.88) & 15.99 \\
        ecbdl14-1.2mill-90 & 960,000 & 90 & (98.01; 1.99) & 49.29 \\
        ecbdl14-10mill-90 & 9,600,000 & 90 & (98.00; 2.00) & 48.94 \\
        kddcup\_normal\_vs\_DOS & 1,942,816 & 41 & (79.96; 20.04) & 3.99 \\
        kddcup\_DOS\_vs\_R2L & 3,107,709 & 41 & (99.97; 0.03) & 3,475.18 \\
		\bottomrule 
	\end{tabular}
\end{table}

All datasets have been partitioned using a 5 fold cross-validation scheme. This means that all datasets have been partitioned in 5 folds, with 80\% of instances devoted to training, and the rest 20\% for testing. The results provided are the average of running the algorithms with the five folds per dataset.

We have carried out a comparison of SD\_DeTE methodology against three classification methods: Spark's MLlib distributed implementation of decision trees, Random Forest, and PCARDE algorithm, a data preprocessing based ensemble present in Spark's community repository Spark Packages~\cite{GARCIAGIL2018}. For balancing the data when those classifiers are used, we have employed the most popular and widely used data balancing methods: RUS, ROS and SMOTE. SMOTE is the state-of-the-art in performance, while ROS combined with Random Forest constitutes the current state-of-the-art in imbalanced Big Data scenario~\cite{Fernandez2017, juez2021experimental}.

For SMOTE algorithm, an implementation available in the Spark Packages repository has been used: SMOTE\_BD~\cite{basgall2018smote}. The parameters used for the data preprocessing algorithms and the different classifiers are described in Table~\ref{tab:params}. Since ensembles correct errors across many base classifiers, we have chosen to increase the depth of the decision tree in SD\_DeTE methodology for a better discrimination between both minority and majority classes. ROS and SMOTE\_BD have been configured to balance the dataset to an $IR = 1$. 

\begin{table}[htb!]
	\centering
	\caption{Parameter settings for the data preprocessing and classification algorithms}
	\label{tab:params}
		\begin{tabular}{ll}
			\toprule
			Algorithm & Parameters \\
			\midrule
			ROS\_BD & ir = 1 \\
			
            SMOTE\_BD & k = 5, distance = ``euclidean'', ir = 1\\
            
            Decision Tree & impurity = ``gini'', maxDepth = 5, maxBins = 32 \\
            
            Random Forest & nTrees = 200, impurity = ``gini'', maxDepth = 4 maxBins = 32 \\
            
            PCARDE & nTrees = 10, bins = 5 \\
			
			SD\_DeTE & bins = 5, trees = 10, maxClust = 10, treeDepth = 10\\
			\bottomrule 
	\end{tabular}
\end{table}

As stated earlier, when dealing with imbalanced data it is crucial to choose the right performance metric. Accuracy is not useful in imbalanced datasets, because we can achieve great accuracy by just classifying correctly the majority class, while the minority class is ignored. For this reason, we have selected the two most widely used metrics for imbalanced classification: GM and AUC.

The experimentation has been carried out in a cluster composed of 11 computing nodes and one master node. The computing nodes have the following hardware characteristics: 2 x Intel Core i7-4930K, 6 cores per processor, 3.40 GHz, 12 MB cache, 4 TB HDD, 64 GB RAM. Regarding software, we have used the following configuration: Apache Hadoop 2.9.1, Apache Spark 2.2.0, 198 cores (18 cores/node), 638 GB RAM (58 GB/node).

\subsection{Results and Analysis}
\label{sec:results}

In this section, we present the results and an analysis of the performance metrics obtained by the selected methods. We denote with \textit{Baseline} the application of the classifiers without using any imbalanced data treatment technique.

In Table~\ref{tab:gm} we can see the average results for the GM measure using the three classifiers combined with the three data preprocessing strategies, compared with SD\_DeTE methodology. As can be observed, the \textit{Baseline} with no data imbalanced handling often results in a GM value of 0. That value represents that one of the classes (the minority in particular) is being missclassified completely. All classifiers are benefiting from the data balancing done by RUS and ROS. All three classifiers achieve very similar results when using either RUS or ROS. This can be explained by the high data redundancy present in Big Data datasets. SMOTE\_BD is able to achieve an improvement in the GM measure when using PCARDE algorithm as a classifier. SD\_DeTE methodology is the best performing method for almost every tested dataset. On average, SD\_DeTE methodology achieves an improvement of nearly 0.5 points in the GM measure. This shows the good performance of the clustering-based data oversampling of SD\_DeTE methodology.

\begin{landscape}
    \begin{table}[htb!]
    	\centering
    	\caption{Average results for the imbalanced Big Data cases of study using the GM measure. The highest GM value per dataset is stressed in bold.}
    	\label{tab:gm}
        \footnotesize
        \begin{tabular}{l|lll|lll|lll|lll|l}
            \toprule
              Dataset  & \multicolumn{2}{l}{Baseline} & &RUS   &       &       & ROS   &        &        & \multicolumn{2}{l}{SMOTE\_BD}  &      & SD\_DeTE    \\
             & DT    & RF    & PCARDE & DT    & RF    & PCARDE & DT    & RF     & PCARDE  & DT    & RF     & PCARDE  &        \\
            \midrule
            poker0\_vs\_2 & 0.1986 & 0.0000 & 0.0000 & 0.5847 & 0.7086 & 0.5859 & 0.5272 & 0.5455 & 0.5813 & 0.5249 & 0.4846 & 0.6604 & \textbf{0.8274} \\
            poker0\_vs\_3 & 0.1261 & 0.0000 & 0.0000 & 0.5248 & 0.6954 & 0.6574 & 0.6890 & 0.7003 & 0.6423 & 0.5728 & 0.5728 & 0.6983 & \textbf{0.8324} \\
            poker0\_vs\_4 & 0.2383 & 0.0000 & 0.0000 & 0.8427 & 0.8407 & 0.8438 & 0.8468 & 0.8481 & 0.9029 & 0.7773 & 0.7757 & 0.9102 & \textbf{0.9880} \\
            poker0\_vs\_5 & 0.0000 & 0.0000 & 0.7002 & 0.8745 & 0.8530 & 0.9735 & 0.8745 & 0.8743 & 0.9555 & 0.4840 & 0.4582 & \textbf{1.0000} & 0.9974 \\
            poker0\_vs\_6 & 0.0000 & 0.0000 & 0.0000 & 0.6197 & 0.6615 & 0.7250 & 0.5935 & 0.7005 & 0.5860 & 0.6209 & 0.5729 & 0.7415 & \textbf{0.7998} \\
            poker1\_vs\_2 & 0.0367 & 0.0000 & 0.0000 & 0.5993 & 0.5437 & 0.4893 & 0.5600 & 0.5539 & 0.5328 & 0.4136 & 0.3452 & 0.5380 & \textbf{0.6635} \\
            poker1\_vs\_3 & 0.0776 & 0.0000 & 0.0402 & 0.5948 & 0.5981 & 0.5193 & 0.6129 & 0.6204 & 0.5347 & 0.5073 & 0.4543 & 0.5720 & \textbf{0.6396} \\
            poker1\_vs\_4 & 0.0000 & 0.0000 & 0.0000 & 0.7675 & 0.7506 & 0.8620 & 0.7678 & 0.7424 & 0.8789 & 0.6571 & 0.7050 & 0.8375 & \textbf{0.9361} \\
            poker1\_vs\_5 & 0.0000 & 0.0000 & 0.7002 & 0.5423 & 0.7845 & 0.9522 & 0.5833 & 0.6073 & 0.9999 & 0.4649 & 0.4574 & 0.9964 & \textbf{1.0000} \\
            poker1\_vs\_6 & 0.0000 & 0.0000 & 0.0000 & 0.6190 & 0.5673 & 0.6105 & 0.6359 & 0.6269 & 0.5129 & 0.6327 & 0.5611 & 0.6060 & \textbf{0.6576} \\
            rlcp & 0.0874 & 0.0927 & 0.0927 & 0.9310 & 0.9302 & 0.9301 & 0.9299 & 0.9305 & 0.9310 & 0.9306 & 0.9297 & 0.9302 & \textbf{0.9313} \\
            susy\_ir4 & 0.6870 & 0.6187 & 0.6615 & 0.7679 & 0.7651 & 0.7737 & 0.7679 & 0.7647 & 0.7748 & 0.7622 & 0.7654 & 0.7757 & \textbf{0.7824} \\
            susy\_ir8 & 0.5713 & 0.5482 & 0.5690 & 0.7671 & 0.7660 & 0.7737 & 0.7678 & 0.7655 & 0.7738 & 0.7623 & 0.7661 & 0.7746 & \textbf{0.7802} \\
            susy\_ir16 & 0.5162 & 0.5205 & 0.4531 & 0.7667 & 0.7651 & 0.7725 & 0.7661 & 0.7647 & 0.7728 & 0.7627 & 0.7654 & 0.7746 & \textbf{0.7815} \\
            higgs\_ir4 & 0.3498 & 0.0541 & 0.2712 & 0.6584 & 0.6695 & 0.6927 & 0.6613 & 0.6702 & 0.6891 & 0.6446 & 0.6622 & 0.6857 & \textbf{0.7141} \\
            higgs\_ir8 & 0.2398 & 0.0000 & 0.1774 & 0.6612 & 0.6698 & 0.6893 & 0.6630 & 0.6688 & 0.6841 & 0.6479 & 0.6511 & 0.6827 & \textbf{0.7174} \\
            higgs\_ir16 & 0.1368 & 0.0000 & 0.0000 & 0.6575 & 0.6679 & 0.6874 & 0.6600 & 0.6691 & 0.6886 & 0.6512 & 0.6506 & 0.6812 & \textbf{0.7121} \\
            ecbdl14-1.2mill-90 & 0.0143 & 0.0000 & 0.0000 & 0.7006 & 0.7056 & 0.7067 & 0.7001 & 0.7032 & 0.7141 & 0.6662 & 0.6920 & 0.6920 & \textbf{0.7225} \\
            ecbdl14-10mill-90 & 0.0000 & 0.0000 & 0.0000 & 0.6979 & 0.7047 & 0.7073 & 0.6976 & 0.7039 & 0.7082 & 0.6736 & 0.6850 & 0.6885 & \textbf{0.7272} \\
            kddcup\_normal\_vs\_DOS & 0.9998 & 0.9998 & 0.9998 & 0.9996 & 0.9996 & 0.9998 & 0.9996 & 0.9996 & 0.9998 & 0.9997 & 0.9996 & 0.9998 & \textbf{1.0000} \\
            kddcup\_DOS\_vs\_R2L & 0.9756 & 0.9934 & 0.9912 & 0.9976 & 0.9997 & 0.9998 & 0.9934 & \textbf{1.0000} & 0.9978 & 0.0000 & \textbf{1.0000} & 0.9976 & 0.9978 \\
            \midrule
            Average & 0.2502 & 0.1823 & 0.2694 & 0.7226 & 0.7451 & 0.7596 & 0.7285 & 0.7362 & 0.7553 & 0.6265 & 0.6645 & 0.7735 & \textbf{0.8194} \\
            \bottomrule
        \end{tabular}
    \end{table}
\end{landscape}

\begin{landscape}
    \begin{table}[htb!]
    	\centering
    	\caption{Average results for the imbalanced Big Data cases of study using the AUC measure. The highest AUC value per dataset is stressed in bold.}
    	\label{tab:auc}
        \footnotesize
        \begin{tabular}{l|lll|lll|lll|lll|l}
            \toprule
              Dataset  & \multicolumn{2}{l}{Baseline} & &RUS   &       &       & ROS   &        &        & \multicolumn{2}{l}{SMOTE\_BD}  &      & SD\_DeTE    \\
             & DT    & RF    & PCARDE & DT    & RF    & PCARDE & DT    & RF     & PCARDE  & DT    & RF     & PCARDE  &        \\
            \midrule
            poker0\_vs\_2 & 0.5197 & 0.5000 & 0.5000 & 0.5456 & 0.6045 & 0.6152 & 0.6148 & 0.7093 & 0.6145 & 0.5997 & 0.5919 & 0.6653 & \textbf{0.8274} \\
            poker0\_vs\_3 & 0.5080 & 0.5000 & 0.5000 & 0.6946 & 0.7191 & 0.6651 & 0.5733 & 0.7151 & 0.6648 & 0.6174 & 0.6174 & 0.7030 & \textbf{0.8326} \\
            poker0\_vs\_4 & 0.5284 & 0.5000 & 0.5000 & 0.8477 & 0.8500 & 0.9029 & 0.8431 & 0.8440 & 0.8440 & 0.7787 & 0.7785 & 0.9102 & \textbf{0.9880} \\
            poker0\_vs\_5 & 0.5000 & 0.5000 & 0.7451 & 0.8824 & 0.8822 & 0.9565 & 0.8824 & 0.8638 & 0.9738 & 0.5028 & 0.5177 & \textbf{1.0000} & 0.9974 \\
            poker0\_vs\_6 & 0.5000 & 0.5000 & 0.5000 & 0.5935 & 0.7015 & 0.6408 & 0.6920 & 0.7160 & 0.7284 & 0.6928 & 0.6543 & 0.7427 & \textbf{0.8167} \\
            poker1\_vs\_2 & 0.5007 & 0.5000 & 0.5000 & 0.6089 & 0.5887 & 0.5521 & 0.6174 & 0.5872 & 0.5513 & 0.5016 & 0.4937 & 0.5453 & \textbf{0.6647} \\
            poker1\_vs\_3 & 0.5030 & 0.5000 & 0.5008 & 0.6129 & 0.6307 & 0.5763 & 0.6034 & 0.6337 & 0.5701 & 0.5596 & 0.5386 & 0.5734 & \textbf{0.6430} \\
            poker1\_vs\_4 & 0.5000 & 0.5000 & 0.5000 & 0.7751 & 0.7551 & 0.8790 & 0.7678 & 0.7607 & 0.8646 & 0.6647 & 0.7173 & 0.8384 & \textbf{0.9372} \\
            poker1\_vs\_5 & 0.5000 & 0.5000 & 0.7452 & 0.5833 & 0.6073 & 0.9999 & 0.5516 & 0.8068 & 0.9532 & 0.4979 & 0.5226 & 0.9964 & \textbf{1.0000} \\
            poker1\_vs\_6 & 0.5000 & 0.5000 & 0.5000 & 0.6455 & 0.6269 & 0.5960 & 0.6383 & 0.6047 & 0.6200 & 0.6440 & 0.5912 & 0.6380 & \textbf{0.6927} \\
            rlcp & 0.5038 & 0.5043 & 0.5043 & 0.9318 & 0.9322 & 0.9328 & 0.9322 & 0.9319 & 0.9320 & 0.9325 & 0.9315 & 0.9320 & \textbf{0.9327} \\
            susy\_ir4 & 0.7260 & 0.6856 & 0.7115 & 0.7687 & 0.7665 & 0.7768 & 0.7689 & 0.7668 & 0.7763 & 0.7642 & 0.7656 & 0.7769 & \textbf{0.7854} \\
            susy\_ir8 & 0.6603 & 0.6477 & 0.6592 & 0.7688 & 0.7670 & 0.7754 & 0.7679 & 0.7678 & 0.7759 & 0.7645 & 0.7664 & 0.7761 & \textbf{0.7821} \\
            susy\_ir16 & 0.6315 & 0.6333 & 0.6020 & 0.7667 & 0.7662 & 0.7758 & 0.7671 & 0.7664 & 0.7741 & 0.7677 & 0.7656 & 0.7757 & \textbf{0.7838} \\
            higgs\_ir4 & 0.5535 & 0.5014 & 0.5344 & 0.6638 & 0.6703 & 0.6891 & 0.6636 & 0.6695 & 0.6930 & 0.6542 & 0.6640 & 0.6858 & \textbf{0.7142} \\
            higgs\_ir8 & 0.5270 & 0.5000 & 0.5153 & 0.6640 & 0.6689 & 0.6841 & 0.6633 & 0.6699 & 0.6896 & 0.6484 & 0.6542 & 0.6829 & \textbf{0.7174} \\
            higgs\_ir16 & 0.5091 & 0.5000 & 0.5000 & 0.6640 & 0.6692 & 0.6887 & 0.6638 & 0.6680 & 0.6874 & 0.6519 & 0.6541 & 0.6814 & \textbf{0.7122} \\
            ecbdl14-1.2mill-90 & 0.5001 & 0.5000 & 0.5000 & 0.7006 & 0.7034 & 0.7141 & 0.7029 & 0.7056 & 0.7068 & 0.6700 & 0.6939 & 0.6943 & \textbf{0.7236} \\
            ecbdl14-10mill-90 & 0.5000 & 0.5000 & 0.5000 & 0.6977 & 0.7039 & 0.7083 & 0.6979 & 0.7047 & 0.7073 & 0.6799 & 0.6885 & 0.6901 & \textbf{0.7273} \\
            kddcup\_normal\_vs\_DOS & 0.9998 & 0.9998 & 0.9998 & 0.9996 & 0.9996 & 0.9998 & 0.9996 & 0.9996 & 0.9998 & 0.9997 & 0.9996 & 0.9998 & \textbf{1.0000} \\
            kddcup\_DOS\_vs\_R2L & 0.9759 & 0.9934 & 0.9912 & 0.9934 & \textbf{1.0000} & 0.9978 & 0.9976 & 0.9997 & 0.9998 & 0.5000 & \textbf{1.0000} & 0.9976 & 0.9978 \\
            \midrule
            Average & 0.5784 & 0.5698 & 0.5957 & 0.7337 & 0.7435 & 0.7679 & 0.7338 & 0.7567 & 0.7679 & 0.6711 & 0.6955 & 0.7764 & \textbf{0.8227} \\
            \bottomrule
        \end{tabular}
    \end{table}
\end{landscape}

The AUC average results are depicted in Table~\ref{tab:auc}. Again, the \textit{Baseline} with no preprocessing achieves low values of AUC. The first difference when comparing AUC with the GM measure, is that AUC shows a value of 0.5 when a full class is completely missclassified. RUS and ROS methods are producing very similar results in terms of AUC measure. Regarding SMOTE\_BD, as observed with the GM measure, only PCARDE algorithm is able to achieve an AUC improvement with respect to RUS and ROS. The same improvement seen with the GM measure can be seen with the AUC measure for SD\_DeTE methodology. It is the best performing data preprocessing and ensemble method among the different strategies tested.

If we attend to the relation between the IR and the performance of SD\_DeTE methodology, we observe that SD\_DeTE methodology is not affected by the different IR's presence in the tested datasets. SD\_DeTE methodology is a very stable ensemble method, achieving almost the same performance for an increasing IR for the same dataset. This behavior can be seen in Susy and Higgs datasets, which have an IR ranging from 4 up to 16, and both the GM and AUC measures are unaffected by the increasing IR. Moreover, some of the tested datasets have an extremely high IR, such as poker0\_vs\_6 and poker1\_vs\_6 datasets, with an IR of 337.81 and 308.77 respectively. For such datasets, SD\_DeTE methodology is the best performing method, with a difference of more than 5\% better performance.

Results presented have shown the excellent performance of our proposal. For a deeper analysis of the results, we have performed a Bayesian Sign Test in order to analyze if SD\_DeTE methodology is statistically better than the rest of the methods~\cite{10.1007/978-3-319-59650-1_24}. Bayesian Sign Tests obtain a distribution of the differences between two algorithms, and make a decision when 95\% of the distribution is in one of the three regions: left, rope (region of practical equivalence), or right~\cite{benavoli2017time}.

\begin{figure}[H]
	\minipage{0.3\textwidth}
	\includegraphics[width=\linewidth]{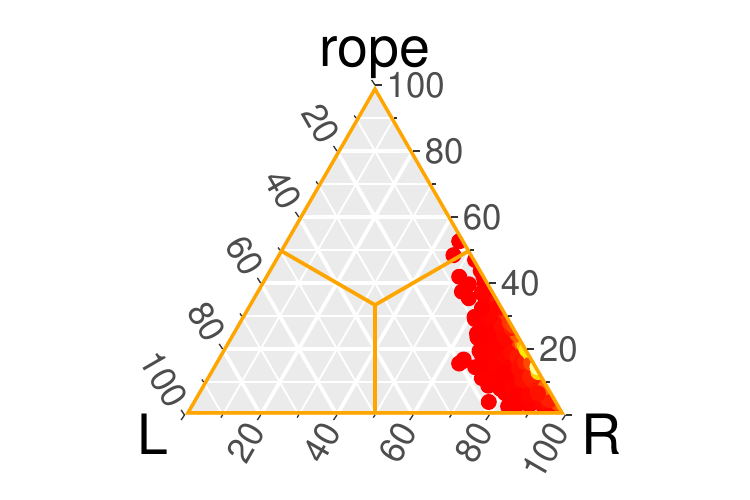}
	\caption*{DT (L) vs SD\_DeTE (R)}
	\endminipage\hfill
	\minipage{0.3\textwidth}
	\includegraphics[width=\linewidth]{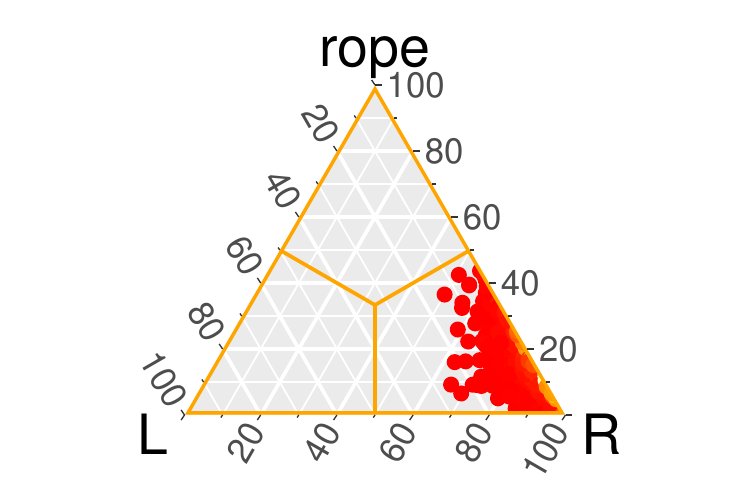}
	\caption*{RF (L) vs SD\_DeTE (R)}
	\endminipage\hfill
	\minipage{0.3\textwidth}
	\includegraphics[width=\linewidth]{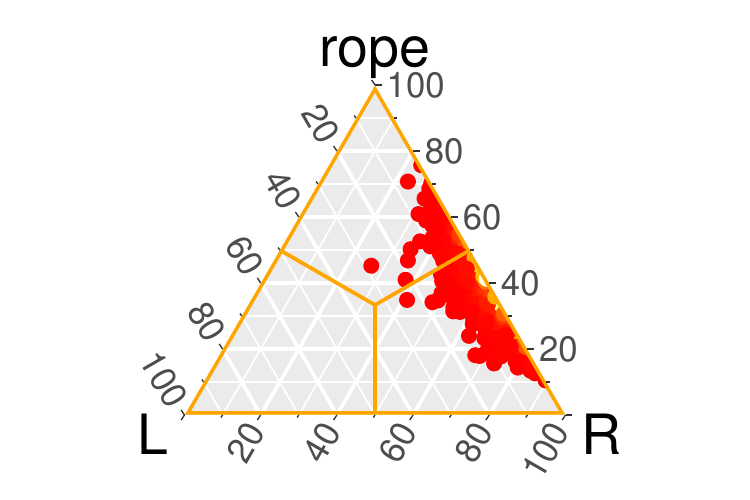}
	\caption*{PCARDE (L) vs SD\_DeTE (R)}
	\endminipage
	\caption{Bayesian Sign Test heatmap of DT, RF and PCARDE best results, against SD\_DeTE methodology for GM measure}
	\label{fig:gm_rope}
\end{figure}

\begin{figure}[H]
	\minipage{0.3\textwidth}
	\includegraphics[width=\linewidth]{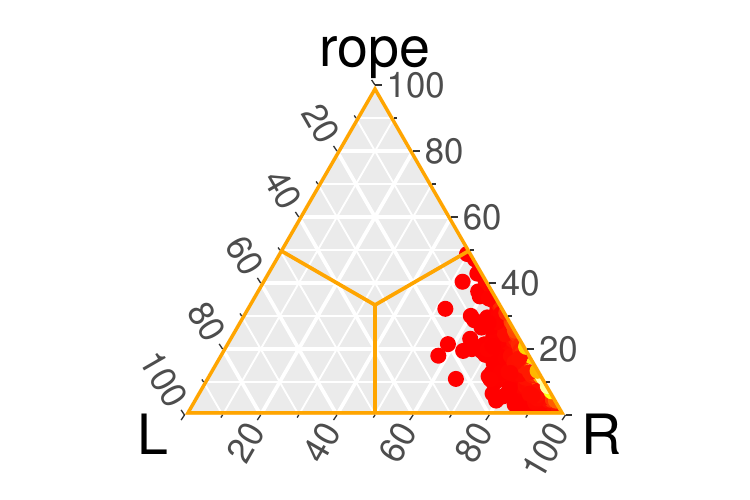}
	\caption*{DT (L) vs SD\_DeTE (R)}
	\endminipage\hfill
	\minipage{0.3\textwidth}
	\includegraphics[width=\linewidth]{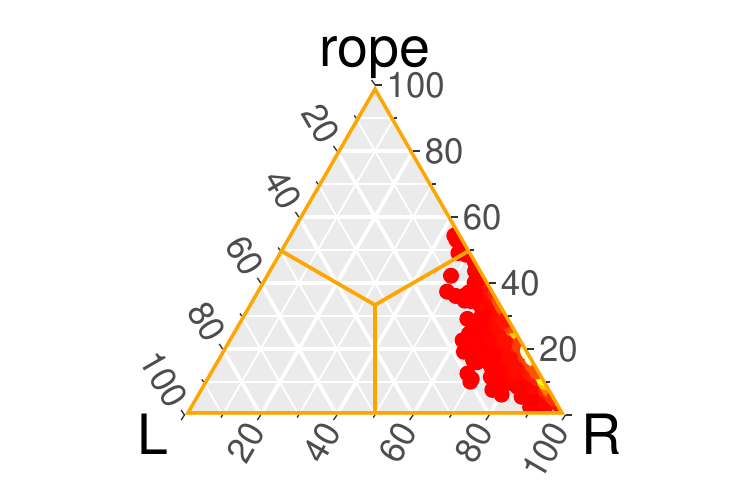}
	\caption*{RF (L) vs SD\_DeTE (R)}
	\endminipage\hfill
	\minipage{0.3\textwidth}
	\includegraphics[width=\linewidth]{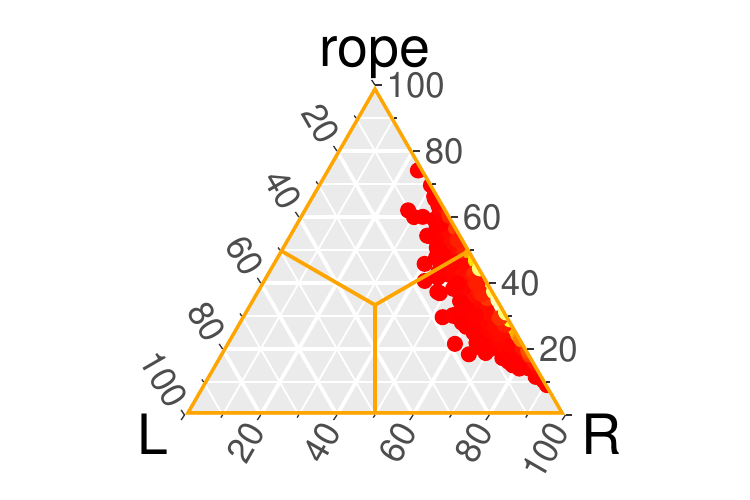}
	\caption*{PCARDE (L) vs SD\_DeTE (R)}
	\endminipage
	\caption{Bayesian Sign Test heatmap of DT, RF and PCARDE best results, against SD\_DeTE methodology for AUC measure}
	\label{fig:auc_rope}
\end{figure}

The Bayesian Sign Test is applied to the mean GM and AUC measures of each dataset. We have selected the best performing scenario for each classification method depending on the measure employed. In Figure~\ref{fig:gm_rope} we can see a comparison of SD\_DeTE methodology against the decision tree with ROS, Random Forest with RUS, and PCARDE algorithm with SMOTE\_BD, all using the GM measure. On the other hand, for AUC measure (showed in Figure~\ref{fig:auc_rope}), the decision tree is combined with RUS, Random Forest with ROS, and PCARDE algorithm with SMOTE\_BD. As we can observe, both GM and AUC Bayesian Sign Tests are showing very similar results. The probability of the difference being to the left is minimal for SD\_DeTE methodology. This means that the Bayesian Sign Test indicating a zero probability for these classification methods to perform better than our proposal.

These results have shown the importance of choosing the correct imbalanced data treatment. SD\_DeTE methodology stands as the best choice for dealing with imbalanced Big Datasets, being able to create an ensemble with efficient distributed algorithms by using Smart Data. SD\_DeTE methodology has achieved statistically the best performance in both GM and AUC for almost every tested dataset, proving its efficiency when dealing with Big Data imbalanced datasets.

\vspace*{-0.5cm}

\subsection{Computing Times}
\label{sec:times}

In order to assess the performance in Big Data scenarios, we shall analyze the computing times for SD\_DeTE methodology and the rest of the methods. In classification tasks, prediction times are more important than learning times, since models are only learned once but used multiple times in prediction. Such times can be seen in Table~\ref{tab:pred_times}. As expected, the decision tree is the fastest in prediction, since it only requires to predict a simple tree. Random Forest also achieve good predictions times, since neither the decision tree nor Random Forest use any data preprocessing techniques when predicting. In spite of this, SD\_DeTE methodology is very competitive in prediction, being less than one second slower than PCARDE. SD\_DeTE is able to predict Big imbalanced Datasets in a short amount of time.

\begin{landscape}
	\begin{table}[htb!]
		\centering
		\caption{Average prediction times (in seconds) for the imbalanced Big Data cases of study.}
		\label{tab:pred_times}
		\footnotesize
		\begin{tabular}{l|lll|lll|lll|lll|l}
			\toprule
			Dataset  & \multicolumn{2}{l}{Baseline} & &RUS   &       &       & ROS   &        &        & \multicolumn{2}{l}{SMOTE\_BD}  &      & SD\_DeTE    \\
			& DT    & RF    & PCARDE & DT    & RF    & PCARDE & DT    & RF     & PCARDE  & DT    & RF     & PCARDE  &        \\
			\midrule
			poker0\_vs\_2 & 0.07 & 1.96 & 3.38 & 0.03 & 1.98 & 2.32 & 0.03 & 1.91 & 2.74 & 0.03 & 1.85 & 2.58 & 3.60 \\
			poker0\_vs\_3 & 0.07 & 1.92 & 3.28 & 0.03 & 1.68 & 2.23 & 0.03 & 1.65 & 2.61 & 0.03 & 1.84 & 2.51 & 3.19 \\
			poker0\_vs\_4 & 0.08 & 1.72 & 3.19 & 0.03 & 1.72 & 2.12 & 0.04 & 1.80 & 2.61 & 0.03 & 1.63 & 2.18 & 3.07 \\
			poker\_0\_vs\_5 & 0.59 & 1.59 & 3.10 & 0.03 & 1.67 & 2.41 & 0.03 & 1.50 & 2.53 & 0.03 & 1.59 & 2.12 & 2.38 \\
			poker0\_vs\_6 & 0.08 & 1.82 & 3.09 & 0.03 & 1.88 & 2.45 & 0.03 & 1.72 & 2.55 & 0.03 & 1.66 & 2.16 & 2.75 \\
			poker1\_vs\_2 & 0.08 & 1.71 & 2.95 & 0.02 & 1.66 & 1.98 & 0.03 & 1.57 & 2.29 & 0.03 & 1.59 & 2.00 & 3.62 \\
			poker1\_vs\_3 & 0.08 & 1.82 & 3.13 & 0.03 & 1.67 & 2.21 & 0.03 & 1.70 & 2.19 & 0.02 & 1.77 & 2.02 & 3.16 \\
			poker1\_vs\_4 & 0.08 & 1.67 & 2.75 & 0.03 & 1.59 & 2.18 & 0.03 & 1.54 & 2.19 & 0.03 & 1.59 & 1.84 & 2.34 \\
			poker1\_vs\_5 & 0.61 & 1.47 & 2.97 & 0.03 & 1.45 & 1.85 & 0.03 & 1.43 & 2.31 & 0.03 & 1.48 & 1.95 & 2.50 \\
			poker1\_vs\_6 & 0.08 & 1.54 & 2.89 & 0.03 & 1.41 & 2.09 & 0.04 & 1.36 & 2.26 & 0.03 & 1.34 & 1.95 & 2.44 \\
			rlcp & 0.14 & 2.30 & 13.28 & 0.04 & 2.28 & 12.29 & 0.04 & 2.30 & 12.37 & 0.04 & 2.26 & 12.64 & 15.35 \\
			susy\_ir4 & 0.08 & 1.37 & 9.05 & 0.04 & 1.40 & 8.31 & 0.04 & 1.48 & 8.14 & 0.04 & 1.54 & 8.18 & 11.34 \\
			susy\_ir8 & 0.10 & 1.07 & 8.18 & 0.04 & 1.17 & 8.30 & 0.05 & 1.09 & 7.79 & 0.04 & 1.11 & 7.96 & 10.77 \\
			susy\_ir16 & 0.10 & 1.14 & 7.62 & 0.04 & 1.25 & 7.22 & 0.04 & 1.20 & 7.22 & 0.04 & 1.23 & 7.61 & 9.94 \\
			higgs\_ir4 & 0.23 & 2.97 & 17.31 & 0.06 & 2.65 & 16.76 & 0.06 & 2.89 & 16.81 & 0.06 & 2.87 & 16.39 & 22.86 \\
			higgs\_ir8 & 0.12 & 2.28 & 16.59 & 0.09 & 2.35 & 15.21 & 0.07 & 2.21 & 15.01 & 0.06 & 2.29 & 15.00 & 21.73 \\
			higgs\_ir16 & 0.24 & 2.45 & 14.71 & 0.26 & 2.43 & 13.80 & 0.06 & 2.43 & 14.19 & 0.06 & 2.54 & 13.57 & 18.96 \\
			ecbdl14-1.2mill-90 & 0.19 & 0.62 & 5.15 & 0.04 & 0.67 & 4.72 & 0.04 & 0.65 & 4.64 & 0.04 & 0.87 & 4.30 & 6.66 \\
			ecbdl14-10mill-90 & 0.21 & 5.24 & 30.81 & 0.05 & 5.56 & 31.42 & 0.06 & 5.57 & 31.17 & 0.06 & 5.66 & 30.87 & 44.36 \\
			kddcup\_normal\_vs\_DOS & 0.21 & 0.99 & 6.44 & 0.04 & 0.86 & 5.99 & 0.04 & 0.95 & 5.99 & 0.04 & 0.94 & 6.15 & 9.01 \\
			kddcup\_DOS\_vs\_R2L & 0.12 & 2.48 & 9.67 & 0.03 & 2.09 & 8.94 & 0.04 & 2.02 & 9.04 & 0.04 & 2.19 & 8.99 & 9.83 \\
			\midrule
			Average & 0.10 & 2.22 & 6.52 & 0.03 & 2.04 & 5.63 & 0.03 & 1.96 & 5.89 & 0.03 & 2.02 & 5.78 & 6.72 \\
			\bottomrule
		\end{tabular}
	\end{table}
\end{landscape}

\subsection{Clustering-based ROS vs ROS}
\label{sec:cro}

We have conducted an additional experiment for analyzing the performance of our proposed clustering-based ROS. Tables~\ref{tab:cro_gm} and~\ref{tab:cro_auc} show the results of SD\_DeTE methodology against SD\_DeTE methodology without using the clustering-based ROS (only performing ROS), using both GM and AUC measures.

\begin{table}[H]
	\centering
	\caption{Average results for the imbalanced Big Data cases of study using the GM measure. The highest GM value per dataset is stressed in bold.}
	\label{tab:cro_gm}
	\begin{tabular}{l|l|l}
		\toprule
		Dataset                 & SD\_DeTE w/o C-ROS & SD\_DeTE      \\
		&           &                                 		           \\
		\midrule
		poker0\_vs\_2           & 0.7438             & \textbf{0.8274} \\
		poker0\_vs\_3           & 0.8185             & \textbf{0.8324} \\
		poker0\_vs\_4           & 0.9808             & \textbf{0.9880} \\
		poker0\_vs\_5           & \textbf{0.9977}             & 0.9974 \\
		poker0\_vs\_6           & 0.7426             & \textbf{0.7998} \\
		poker1\_vs\_2           & 0.6138             & \textbf{0.6635} \\
		poker1\_vs\_3           & 0.6281             & \textbf{0.6396} \\
		poker1\_vs\_4           & \textbf{0.9390}             & 0.9361 \\
		poker1\_vs\_5           & 1.0000             & \textbf{1.0000} \\
		poker1\_vs\_6           & 0.6394             & \textbf{0.6576} \\
		rlcp                    & 0.9312             & \textbf{0.9313} \\
		susy\_ir4               & 0.7815             & \textbf{0.7824} \\
		susy\_ir8               & \textbf{0.7816}             & 0.7802 \\
		susy\_ir16              & 0.7813             & \textbf{0.7815} \\
		higgs\_ir4              & 0.7139             & \textbf{0.7141} \\
		higgs\_ir8              & 0.7146             & \textbf{0.7174} \\
		higgs\_ir16             & 0.7120             & \textbf{0.7121} \\
		ecbdl14-1.2mill-90      & 0.7220             & \textbf{0.7225} \\
		ecbdl14-10mill-90       & 0.7267             & \textbf{0.7272} \\
		kddcup\_normal\_vs\_DOS & \textbf{1.0000}    & \textbf{1.0000} \\
		kddcup\_DOS\_vs\_R2L    & 0.9977             & \textbf{0.9978} \\
		\midrule
		Average                 & 0.8079             & \textbf{0.8194} \\
		\bottomrule
	\end{tabular}
\end{table}

SD\_DeTE methodology, using the proposed clustering-based ROS, achieves the best results overall. On average, it is improving the performance in both GM and AUC measures by 1 full point. There are datasets where this difference is even more noticeable, such as poker0\_vs\_2, in which the difference increases to more than 8 points. This shows the excellent performance of the proposed clustering-based ROS in the SD\_DeTE methodology.

\begin{table}[H]
	\centering
	\caption{Average results for the imbalanced Big Data cases of study using the AUC measure. The highest AUC value per dataset is stressed in bold.}
	\label{tab:cro_auc}
	\begin{tabular}{l|l|l}
		\toprule
		Dataset                 & SD\_DeTE w/o C-ROS & SD\_DeTE      \\
		&           &                                 		 		   \\
		\midrule
		poker0\_vs\_2           & 0.7438             & \textbf{0.8274} \\
		poker0\_vs\_3           & 0.8185             & \textbf{0.8326} \\
		poker0\_vs\_4           & 0.9809             & \textbf{0.9880} \\
		poker0\_vs\_5           & \textbf{0.9977}    & 0.9974          \\
		poker0\_vs\_6           & 0.7680             & \textbf{0.8167} \\
		poker1\_vs\_2           & 0.6195             & \textbf{0.6647} \\
		poker1\_vs\_3           & 0.6303             & \textbf{0.6430} \\
		poker1\_vs\_4           & \textbf{0.9404}   & 0.9372          \\
		poker1\_vs\_5           & 1.0000             & \textbf{1.0000} \\
		poker1\_vs\_6           & 0.6817             & \textbf{0.6927} \\
		rlcp                    & 0.9326             & \textbf{0.9327} \\
		susy\_ir4               & 0.7846             & \textbf{0.7854} \\
		susy\_ir8               & \textbf{0.7825}    & 0.7821          \\
		susy\_ir16              & 0.7824             & \textbf{0.7838} \\
		higgs\_ir4              & 0.7140             & \textbf{0.7142} \\
		higgs\_ir8              & 0.7156             & \textbf{0.7174} \\
		higgs\_ir16             & 0.7111             & \textbf{0.7122} \\
		ecbdl14-1.2mill-90      & 0.7228             & \textbf{0.7236} \\
		ecbdl14-10mill-90       & 0.7267             & \textbf{0.7273} \\
		kddcup\_normal\_vs\_DOS & \textbf{1.0000}    & \textbf{1.0000} \\
		kddcup\_DOS\_vs\_R2L    & 0.9977             & \textbf{0.9978} \\
		\midrule
		Average                 & 0.8119             & \textbf{0.8227} \\
		\bottomrule
	\end{tabular}
\end{table}

\section{Conclusions}
\label{sec:conclusions}

In this paper, we have proposed a novel Smart Data driven Decision Trees Ensemble methodology for addressing the imbalanced classification problem in Big Data domains, namely SD\_DeTE methodology. SD\_DeTE methodology makes use of the combination of different data preprocessing methods for improving the quality of the data used in the learning of the ensemble. This quality data is able to produce an ensemble composed of efficient distributed algorithms. SD\_DeTE methodology uses RD and PCA for achieving diversity in the Smart Data sets for the ensemble process, plus a novel combination of clustering and oversampling with ROS for achieving a balanced and \textit{smart} dataset while adding another level of diversity.

In view of the results, we can conclude that:

\begin{itemize}
	\item The combination of RD and PCA for adding diversity to the ensemble algorithm achieves excellent performance in imbalanced Big Datasets.
	
	\item The proposed addition of hierarchical clustering and ROS for balancing the data has proven to be able to effectively produce balanced datasets, while adding another level of diversity to the ensemble.
	
	\item SD\_DeTE methodology has proven to be able to achieve efficient distributed algorithms using Smart Data, producing an ensemble capable of tackling Big Data imbalanced problems efficiently and effectively.
\end{itemize}

\section*{Funding}

This work is supported by the Spanish National Research Project PID2020-119478GB-I00. D. García-Gil holds a contract co-financed by the European Social Fund and the Administration of the Junta de Andalucía, reference DOC\_01137. The research is also supported by the Swedish Research Council (project number: 2016-05431).

\section*{Conflict of Interest}
The authors declare that they have no conflict of interests.

\section*{Ethical Approval}
This article does not contain any studies with human participants performed by any of the authors.

\bibliography{Paper}

\end{document}